\documentclass[10pt,twocolumn,letterpaper]{article}

\usepackage[pagenumbers]{cvpr} 

\usepackage{graphicx}
\usepackage{amsmath}
\usepackage{amssymb}
\usepackage{booktabs}
\usepackage{dsfont}

\usepackage{sidecap}
\usepackage{arydshln}
\newcommand{\tableCellHeight}{1} 
\newcommand{\tabstyle}[1]{
  \setlength{\tabcolsep}{#1}
  \renewcommand{\arraystretch}{\tableCellHeight}
  \centering
  \small
}

\usepackage{times}
\usepackage{epsfig}

\usepackage{setspace}
\usepackage{bm}
\usepackage{multirow}
\usepackage{pifont}
\usepackage{balance}
\usepackage{makecell}
\usepackage{colortbl}
\usepackage{cuted} 
\usepackage{pifont}
\usepackage{tablefootnote}

\usepackage{multicol}
\usepackage{adjustbox}
\definecolor{mygray}{gray}{.9}
\usepackage{wrapfig}
\usepackage{mdframed}

\usepackage{color}

\usepackage{algorithm}
\usepackage[noend]{algpseudocode}

\def\paperID{619} 
\def\confName{CVPR}
\def\confYear{2025}

\definecolor{c1}{RGB}{130,60,120}
\definecolor{c2}{RGB}{223,139,111}
\definecolor{c3}{RGB}{144,59,28}
\usepackage{xcolor}
\usepackage[normalem]{ulem} 
\newcommand\hl{\bgroup\markoverwith
  {\textcolor{yellow}{\rule[-.5ex]{2pt}{2.5ex}}}\ULon}
\definecolor{cvprblue}{rgb}{0.21,0.49,0.74}

\usepackage[pagebackref,breaklinks,colorlinks,allcolors=cvprblue]{hyperref}

\usepackage[capitalize]{cleveref}
\crefname{section}{Sec.}{Secs.}
\Crefname{section}{Section}{Sections}
\Crefname{table}{Table}{Tables}
\crefname{table}{Tab.}{Tabs.}

\begin{document}

\title{POPEN: Preference-Based Optimization and Ensemble for\\LVLM-Based Reasoning Segmentation}
\author{
    Lanyun Zhu\textsuperscript{\rm1}~~~
    Tianrun Chen\textsuperscript{\rm3}~~~
    Qianxiong Xu\textsuperscript{\rm4}~~~
    Xuanyi Liu\textsuperscript{\rm5}\\
    Deyi Ji\textsuperscript{\rm2}~~~
    Haiyang Wu\textsuperscript{\rm2}~~~
    De Wen Soh\textsuperscript{\rm1}~~~ 
    Jun Liu\textsuperscript{\rm6}\\
    \textsuperscript{\rm1}{\normalsize Singapore University of Technology and Design}~~~
    \textsuperscript{\rm2}{\normalsize Tencent}~~~
    \textsuperscript{\rm3}{\normalsize  Zhejiang University}\\
    \textsuperscript{\rm4}{\normalsize  Nanyang Technological University}~~~
    \textsuperscript{\rm5}{\normalsize  Peking University}~~~
    \textsuperscript{\rm6}{\normalsize Lancaster University} 
}

\maketitle

\begin{abstract}
Existing LVLM-based reasoning segmentation methods often suffer from imprecise segmentation results and hallucinations in their text responses. This paper introduces POPEN, a novel framework designed to address these issues and achieve improved results. POPEN includes a preference-based optimization method to finetune the LVLM, aligning it more closely with human preferences and thereby generating better text responses and segmentation results. Additionally, POPEN introduces a preference-based ensemble method for inference, which integrates multiple outputs from the LVLM using a preference-score-based attention mechanism for refinement. To better adapt to the segmentation task, we incorporate several task-specific designs in our POPEN framework, including a new approach for collecting segmentation preference data with a curriculum learning mechanism, and a novel preference optimization loss to refine the segmentation capability of the LVLM. Experiments demonstrate that our method achieves state-of-the-art performance in reasoning segmentation, exhibiting minimal hallucination in text responses and the highest segmentation accuracy compared to previous advanced methods like LISA and PixelLM. Project page is \href{https://lanyunzhu.site/POPEN/}{here}.
\end{abstract}
\vspace{-1\baselineskip}

\vspace{-0.25\baselineskip}
\section{Introduction}
\begin{figure}[t]
    \centering
    \includegraphics[width=0.9\linewidth]{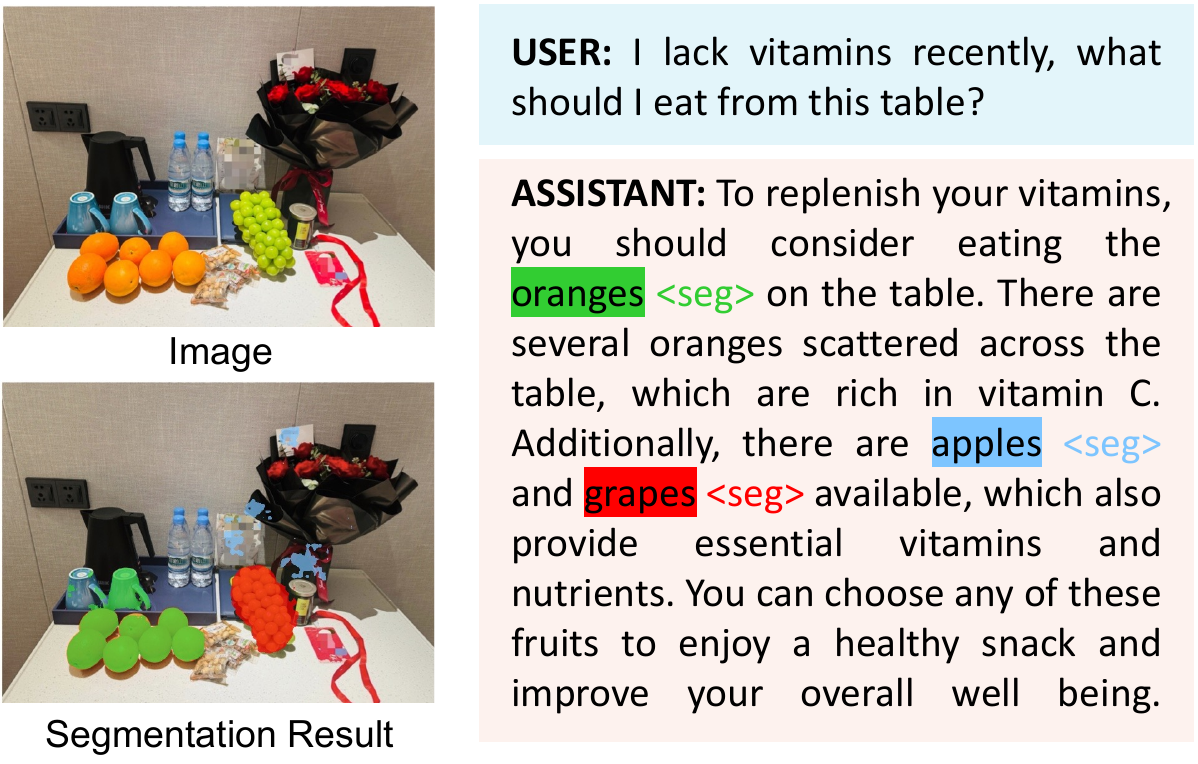}
    \caption{\textbf{An example of hallucination in text responses and inaccurate segmentation results in existing LVLM-based reasoning segmentation methods.} In this example, the LVLM generates the non-existent apple in the text response. The segmentation results show rough edges (grapes) or incorrect localization (misidentifying part of the area belonging to the cup as an orange).}
    \vspace{-0.5\baselineskip}
    \label{intro_fig}
\end{figure}

\vspace{-0.25\baselineskip}
Image segmentation is an important and fundamental task in computer vision that aims to classify each pixel in an image. Traditional methods in this field are typically constrained to segmenting only clearly indicated objects or categories. To overcome this limitation, recent studies, such as LISA \cite{lai2024lisa} and PixelLM \cite{ren2024pixellm}, have leveraged large vision-language models (LVLMs) to enhance the language comprehension capabilities of segmentation algorithms, enabling segmentation to be performed based on more complex human instructions. For example, given the instruction ``I lack vitamins recently, what should I eat from this table?", the model can generate a text response and segment the vegetables and fruits in the image, as shown in Figure \ref{intro_fig}.

While these methods have achieved some success, as shown in Figure \ref{intro_fig}, their performance is still constrained by two severe issues. Firstly, the LVLM frequently generates text responses unrelated to the image content, a problem known as hallucination. This issue can lead to the incorrect generation and segmentation of non-existent objects within the image, such as the “apple” in Figure \ref{intro_fig}. Secondly, the segmentation accuracy is often suboptimal, with coarse results at object boundaries and even incorrect localization of the target objects. One possible reason for this issue is that the SFT-trained LVLM has not yet developed sufficient capability to generate highly refined segmentation features. These two challenges reveal the lack of robustness and effectiveness in current LVLM-based reasoning segmentation models, underscoring the need for a more effective training paradigm to further enhance the LVLM’s segmentation capabilities and mitigate the issue of hallucinations.

In this paper, we propose a novel framework named POPEN, which effectively addresses the aforementioned issues and achieves significantly improved performance. Our core idea is to align the model’s outputs with human preference through reinforcement learning, inspired by the success of preference optimization methods \cite{christiano2017deep, rafailov2024direct} in improving language models. This method refines the LVLM by training it to differentiate between high-quality, human-preferred responses and less desirable ones, thus producing better results with reduced hallucinations and enhanced segmentation precision. We find that directly using classical preference optimization methods from NLP, such as DPO \cite{rafailov2024direct}, is unsuitable for the reasoning segmentation task, as these methods focus solely on optimizing the quality of the text response but not the accuracy of segmentation results. To address this limitation, we propose a novel preference optimization mechanism specially designed for the segmentation task, with task-tailored designs in both \textit{preference data collection} and \textit{preference optimization loss}. To be specific, we propose a noise-filling method to collect segmentation preference data, along with a curriculum learning mechanism that collects different types of data at different stages to enhance optimization effectiveness. Moreover, a novel loss for segmentation preference optimization is also introduced, addressing the issue that the standard DPO loss is unsuitable for this task due to the infeasibility of calculating the likelihood of the LVLM generating a segmentation embedding. By combining this novel preference optimization method for \textit{segmentation} with another one for \textit{text responses}, our framework is capable of mitigating both the hallucination in text and the inaccuracy of segmentation results, as mentioned in the previous paragraph.

Moreover, to further improve the quality of the text response and segmentation result, we also propose a preference-based ensemble method that integrates multiple different outputs from the LVLM for refinement. During this process, a preference score is computed to adjust the LVLM’s attention, allowing outputs with higher reliability to receive more focus during integration. By combining the proposed \textbf{p}reference-based \textbf{op}timization for finetuning and preference-based \textbf{en}semble for inference, our POPEN demonstrates outstanding performance on the LVLM-based reasoning segmentation task. Experiments on multiple datasets show that POPEN achieves state-of-the-art (SOTA) performance, with significant advantages over previous advanced methods such as LISA and PixelLM. 

In conclusion, the main contributions of our work are as follows: (1) We propose the first preference-based optimization method specifically designed for the reasoning segmentation task, effectively reducing hallucinations and improving segmentation accuracy. (2) We introduce a preference-based ensemble method for multi-output integration, improving the model’s robustness. (3) By integrating the preference-based optimization and ensemble methods, our POPEN achieves SOTA results, as demonstrated by extensive experiments on several benchmarks.

\vspace{-0.5\baselineskip}
\section{Related Work}
\vspace{-0.5\baselineskip}
\noindent \textbf{LVLM-based Image Segmentation.} Image segmentation is a fundamental task in computer vision, and it has achieved significant progress in the era of deep learning \cite{deeplabv1, deeplabv3, cheng2022masked, zhu2023continual, wang2023fvp, zhu2024addressing, xu2024hybrid, unet, zhu2025not, ji2024discrete, kirillov2023segment, zhu2021learning, ji2022structural, ji2025structural, ji2023ultra}. Some recent works \cite{lai2024lisa, ren2024pixellm, wang2023visionllm, zhu2024llafs, zhang2024groundhog, chen2024sam4mllm, yan2024visa, zhang2025psalm, rasheed2024glamm} leverage large vision-language models (LVLMs) \cite{liu2023llava, zhu2023minigpt, li2023blip, zhu2024ibd} to enhance the language comprehension capabilities of segmentation algorithms. For example, LISA \cite{lai2024lisa} proposes the first framework that uses an LVLM followed by a SAM-based decoder for reasoning segmentation. GSVA \cite{xia2024gsva} addresses the shortcomings of LISA by employing multiple [SEG] tokens for multi-target segmentation and a [REJ] token to reject empty targets. LLaFS \cite{zhu2024llafs}, based on VisionLLM \cite{wang2023visionllm}, introduces a novel LVLM-based framework for few-shot segmentation that incorporates a fine-grained instruction and a pseudo-sample-based training method. PixelLM \cite{ren2024pixellm} proposes an improved segmentation feature extraction method and a stronger but more lightweight decoder, achieving both better performance and reduced computational cost. However, these methods often suffer from significant hallucinations and imprecise segmentation. This work introduces a novel preference-based optimization and ensemble method to address these issues and achieves improved performance.

\noindent \textbf{Learning from Humane Feedback.} Recent works have explored aligning large language models with human preferences by learning from human feedback. RLHF \cite{christiano2017deep} proposes the pioneering framework in this field using the proximal policy optimization (PPO) algorithm, but its additional reward model and complex reinforcement learning framework increase the difficulty of model training. Direct preference optimization (DPO) \cite{rafailov2024direct} and its extensions \cite{amini2024direct, song2024preference, liustatistical} simplify the RLHF approach by omitting the reward model, significantly reducing computational and storage requirements. Beyond NLP applications, DPO has also been extended to multimodal and computer vision domains such as LVLMs \cite{yu2024rlhf, zhang2024direct} and diffusion-based generation \cite{wallace2024diffusion, zhang2024itercomp}. However, to our knowledge, no human feedback learning method has been specifically designed for LVLM-based reasoning segmentation. Our work constructs a novel framework by proposing the first preference optimization method tailored for this task, incorporating unique designs that can enhance both the text responses and segmentation results. Additionally, we propose a novel preference-based ensemble method that further elevates performance, marking a significant advancement in this field.

\begin{figure*}[t]
    \centering
    \includegraphics[width=1\linewidth]{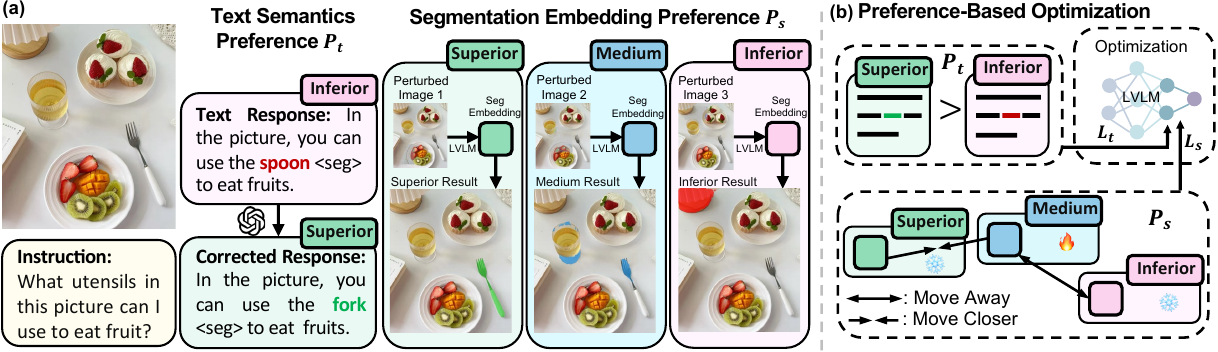}
    \caption{Illustration of \textbf{(a) preference data collection} and \textbf{(b) preference optimization method} in our POPEN framework.}
    \vspace{-0.5\baselineskip}
    \label{main_fig}
\end{figure*}

\vspace{-0.5\baselineskip}
\section{Method}
\vspace{-0.5\baselineskip}
\subsection{Preliminaries and Overview}
\vspace{-0.5\baselineskip}
Existing LVLM-based methods \cite{lai2024lisa, ren2024pixellm} are typically consisted of an LVLM-based encoder followed by a segmentation decoder. The LVLM receives the input image $I$ and instruction $x$ to generate a sequence of text tokens, and a segmentation embedding $f$ associated with each special token \texttt{<seg>} in the text response, for example, the codebook feature $C_{seg}$ in PixelLM \cite{ren2024pixellm}, is extracted and fed into the decoder for segmentation. As detailed in the introduction, these LVLM-based reasoning segmentation methods often suffer from significant hallucinations and imprecise segmentation. To mitigate these issues, this paper introduces a novel framework named POPEN to achieve more effective LVLM-based segmentation by leveraging preference data. Specifically, based on PixelLM as the basic model structure, POPEN employs a segmentation-tailored \textbf{p}reference-based \textbf{op}timization method to improve the reliability of the model’s outputs, and introduces a preference-based \textbf{en}semble framework to integrate information from multiple outputs and further enhance performance. In the following Sec.\ref{text_optim} and Sec.\ref{text_preference_ensemble}, we introduce these two components of our POPEN, respectively.

\subsection{Preference-Based Optimization}\label{text_optim}
We first propose a method to finetune the LVLM using preference data to enhance model performance. Each preference data is basically formatted as $\{I, x, y_{w}, y_{l}\}$, where $I$ and $x$ represent the input image and language instruction respectively; $y_{w}$ and $y_{l}$ refer to two responses, with $y_{w}$ been identified as more aligned with human preferences, and $y_{l}$ as less aligned. The objective is to train the LVLM to differentiate between high-quality responses $y_{w}$ preferred by humans and inferior ones $y_{l}$, thus producing better results with reduced hallucinations and enhanced precision. The first challenge in implementing such a finetuning framework is to collect effective preference data. Given that LVLM-based segmentation methods need to simultaneously address the quality of the text response and the accuracy of the segmentation result, we design a task-specific method to collect two types of data: text semantics preference $\mathcal{P}_{t}$ and segmentation embedding preference $\mathcal{P}_{s}$ as follows:

\noindent \textbf{Text Semantics Preference. }For text semantics preference $\mathcal{P}_{t}$, which focuses on the text component of the response, we employ a classical method proposed in \cite{yu2024rlhf} for its generation. Specifically, for each image-instruction pair $\{I, x\}$ in the MUSE \cite{ren2024pixellm} dataset, we first prompt the SFT-trained LVLM to generate a response $y$ in which segmentation is indicated by the $\texttt{<seg>}$ text token. We then use ChatGPT \footnote{Please see Supp for ChatGPT prompt used to correct the errors in $y$.} to refine $y$ by modifying, adding, or deleting certain words or sentences in $y$, thus generating a corrected response $y_{c}$ with fewer errors and a set $\mathcal{P}_{t}=\{I, x, y, y_{c}, L_{y}, L_{y_{c}}\}$, where $L_{y}$ and $L_{y_{c}}$ refer to two lists which respectively include the position indexes of tokens in $y$ and $y_{c}$ that are different from each other. To enrich the dataset, for some of LVLM's responses that contain only few errors, we instruct ChatGPT to intentionally introduce errors into the ground truth response $y_{g}$ to formulate $y$. Please see Supp for more details.

\noindent \textbf{Segmentation Embedding Preference. }For $\mathcal{P}_{s}$ that focuses on the segmentation embedding $f$ extracted from the LVLM for decoder input (in our method, the codebook feature $C_{seg}$ in the PixelLM network), it is challenging to obtain the preference using the same method as $\mathcal{P}_{t}$, since the implicit embeddings $f$ are difficult to be directly corrected as we did with the text response in $\mathcal{P}_{t}$. To address this, we propose an alternative approach that induces the model to output different $f$ with varying segmentation performance. Specifically, for each pair $\{I, x\}$ whose ground truth response $y_{g}$ contains $N$ target segmentation tokens \texttt{<seg>}, we introduce three different random Gaussian noises to three random rectangular regions in $I$. The model then processes these three perturbed images $\{\hat{I}^{i}\}_{i=1}^{3}$ along with the instruction $x$ to respectively generate three sets of segmentation embeddings $\{\{f^{n, i}\}_{n=1}^{N}\}_{i=1}^{3}$ and their corresponding segmentation masks $\{\{M^{n, i}\}_{n=1}^{N}\}_{i=1}^{3}$. Our empirical observations indicate that the variation in noise can lead to noticeable differences in segmentation outcomes. Therefore, we define the preference data as $\mathcal{P}_{s} = \{\{\hat{I}^{i}\}_{i=1}^{3}, x, \{\{f^{n, i}\}_{n=1}^{N}\}_{i=1}^{3}, \{\{M^{n, i}\}_{n=1}^{N}\}_{i=1}^{3}, \{L_{s}^{n}\}_{n=1}^{N}\}$, where each $L_{s}^{n}$ refers to an index list sorted by segmentation performance, for example, $L_{s}^{n} = [3, 1, 2]$ if $M^{n, 3}$ surpasses $M^{n, 1}$ and $M^{n, 1}$ surpasses $M^{n, 2}$.

\begin{algorithm}[t]
\footnotesize
\captionsetup{font=footnotesize}
\caption{Algorithm of collecting segmentation embedding preference data $\mathcal{P}_{s}$ from an image-instruction pair.}
\label{alg}
\begin{algorithmic}
\State \textbf{Input:} image $I$, instruction $x$, gt masks $\{M_{g}^{n}\}_{n=1}^{N}$, model $\pi$, SAM $S$ 
\State Generate $\{M_{s}^{j}\}_{j=1}^{N_{s}}$ from $I$ using $S$ 
\While{True}
\For{$i$ \textbf{in} $1,2,...,N_{p}$}
\State Generate a random noise $\mathcal{N}$ and a perturbed image 
$\hat{I}^{i}=I+\mathcal{N}$
\State Generate $\{M^{n, i}\}_{n=1}^{N}$ from \{$\hat{I}^{i}, x$\} using $\pi$
\State Compute $s^{i}$ (Eq.\ref{eq_select_score}) and boundary IoU $b^{i}$ for $\hat{I}^{i}$
\EndFor
\If{1st half of finetuning and ${\rm Min}\;s^{i} < 0$ and ${\rm Max}\; s^{i} > 0.8$}
\State Select three $\hat{I}^{i}$ with the highest, median, and lowest $s^{i}$ into $\mathcal{P}_{s}$ 
\State \textbf{break}
\ElsIf{2nd half of finetuning}
\State From $\hat{I}^{i}$ with the top5 highest $s^{i}$, select three $\hat{I}^{i}$ with the highest, 
\State median, and lowest $b^{i}$ into $\mathcal{P}_{s}$
\State \textbf{break}
\EndIf
\EndWhile
\State \textbf{Return:} $\mathcal{P}_{s}$
\end{algorithmic}
\end{algorithm}

\noindent \textbf{Curriculum Collection for $\mathcal{P}_{s}$. }We find that finetuning on $\mathcal{P}_{s}$ obtained through the aforementioned method fails to yield satisfactory improvement. One possible reason, based on our empirical observation, could be that many $\{M^{n,i}\}_{i=1}^{3}$ in the fully-randomly-generated $\mathcal{P}_{s}$ only exhibit differences in the object boundary regions. Consequently, using them for preference-based finetuning may not effectively mitigate segmentation errors outside the boundaries, such as the wrong localization of target objects, which is observed to be a common issue in inference and often has a significant impact on validation accuracy. Previous works \cite{rahaman2019spectral, saxe2019mathematical} have found that deep models typically develop general capabilities such as object localization in the early stages of training and subsequently acquire more refined skills like boundary delineation during later stages. Inspired by this, we propose a curriculum collection mechanism, where different types of $\mathcal{P}_{s}$ are collected and employed in different finetuning stages, allowing the model to first optimize fundamental segmentation skills for target localization, and then improve the precision of boundary processing for further refinement. Specifically, for each pair $\{I, x\}$ with $N$ segmentation targets predicted in its response, we first generate $N_{p}$ perturbed images $\{\hat{I}^{i}\}_{i=1}^{N_{p}}$ by adding different noises to $I$ using the method described in the previous section. We then process $I$ through the Segment Anything Model (SAM) \cite{kirillov2023segment} to produce a set of class-agnostic object masks $\{M_{s}^{j}\}_{j=1}^{N_{s}}$. In the first half of finetuning, our primary focus is on correcting the model’s target localization errors. For this, we compute a score $s^{i}$ for each $\hat{I}^{i}$ by:
 \begin{equation} \label{eq_select_score}
    s^{i} = \frac{1}{N}\sum_{n=1}^{N}\left({\rm IoU}\left(M^{n,i}, M^{n}_{g}\right) - \underset{M_{s}^{j}\notin M^{n}_{g}}{\rm Max}{\rm IoU}\left(M^{n,i}, {M}_{s}^{j}\right)\right),
\end{equation}
where $M^{n,i} \in \{M^{n,i}\}_{n=1}^{N}$ is the $n$-th segmentation mask generated by the LVLM-based model with input $\{\hat{I}^{i}, x\}$, $M^{n}_{g}$ is the ground truth mask for $M^{n,i}$, and $M_{s}^{j}\notin M^{n}_{g}$ refers to $M_{s}^{j} \in \{M_{s}^{j}\}_{j=1}^{N_{s}}$ not corresponding to $M^{n}_{g}$. A lower $s^{i}$ indicates that $\{M^{n,i}\}_{n=1}^{N}$ has a lower overlap with the ground truth $\{M^{n}_{g}\}_{n=1}^{N}$ but higher overlap with other objects. To obtain preference data, we choose three $\hat{I}^{i}$ with the highest, lowest, and medium $s^{i}$ to construct $\mathcal{P}_{s}$. This set $\mathcal{P}_{s}$ thus includes output masks with varying degrees of target localization accuracy, and it is employed to finetune the LVLM for mitigating localization errors. Note that a mechanism to conditionally regenerate perturbed images is employed to ensure that $\mathcal{P}_{s}$ contains sufficiently high-$s$ and low-$s$ samples. Please see Alg.\ref{alg} for details. In the second half of the finetuning process, we shift our focus to optimizing segmentation boundary details when the localization is nearly accurate. To achieve this, from $\hat{I}^{i}$ with the top 5 highest $s^{i}$, we select those with the highest, lowest, and median boundary IoUs to construct $\mathcal{P}_{s}$. 
This dual-phase preference collection enables our method to sequentially optimize the model’s fundamental (target localization) and advanced (boundary refinement) segmentation capabilities in a curriculum learning manner. Experiments presented in Table \ref{table_ablation_collection} demonstrate the effectiveness of this novel approach.

\noindent \textbf{Preference Optimization.}\label{text_optimization} We employ the aforementioned method to construct $\mathcal{P}_{t}$ and the two-phase $\mathcal{P}_{s}$ from all image-instruction pairs in the MUSE \cite{ren2024pixellm} dataset. The next challenge is how to leverage this preference data to finetune the LVLM effectively to mitigate hallucinations and improve segmentation accuracy. A classical method in NLP for utilizing preference data is RLHF \cite{christiano2017deep}, which is effective but typically requires an additional reward model and a reinforcement learning mechanism that are complex to optimize. The recently proposed DPO \cite{rafailov2024direct} simplifies the RLHF framework by eliminating the reward model and directly employing the LVLM itself to compute the reward. Specifically, for the text semantics preference data $\mathcal{P}_{t}=\{I, x, y, y_{c}, L_{y}, L_{y_{c}}\}$, the DPO loss is formulated as:
\begin{equation}\label{eq_dpo}
\begin{aligned}
    \mathcal{L}_{t} &= - \mathbb{E}_{\mathcal{P}_{t}} \left[\log \sigma \left( r(I, x, y_c) - r(I, x, y) \right)\right]\\
    &=-\mathbb{E}_{\mathcal{P}_{t}}[\log \sigma (\beta_{t} \log \frac{\pi_{\theta}(y_c | I,x)}{\pi_{\text{ref}}(y_c | I,x)} - \beta_{t} \log \frac{\pi_{\theta}(y | I,x)}{\pi_{\text{ref}}(y |I,x)})],
\end{aligned}
\end{equation}
where $r$ denotes the reward function, $\beta_{t}$ is a hyperparameter set to 0.5 following \cite{yu2024rlhf}, $\pi_{\theta}$ is the policy LVLM that is continuously updated during finetuning, and $\pi_{\text{ref}}$ is a reference LVLM that is frozen at the initial state of $\pi_{\theta}$. For $\mathcal{P}_{t}$, we follow \cite{yu2024rlhf} to compute $\log \pi(y|I,x)$ in Eq.\ref{eq_dpo} by weighted summing the likelihood of all tokens in \( y \). Formally, 
\begin{equation} \label{eq_logpi}
\begin{aligned}
    \log \pi(y|I, x) = \frac{1}{|y|} &(\sum_{i \notin L_{y}} \log p(y^i | I, x, y^{<i}) \\ 
     + \lambda &\sum_{i \in L_{y}} \log p(y^i | I, x, y^{<i})), 
\end{aligned}
\end{equation}
where $y^{i}$ is the $i$-th token in $y$, $L_{y}$ refers to the position index list for tokens different between $y$ and $y_{c}$. $\lambda=5$ (following \cite{yu2024rlhf}) is a hyperparameter that assigns higher weight to the tokens corrected by ChatGPT, as they are more likely to contain hallucinated content. We use the same method to compute $\log \pi(y_{c}|I,x)$ from $y_{c}$. $\log \pi(y|I,x)$ and $\log \pi(y_{c}|I,x)$ are employed in Eq.\ref{eq_dpo} to compute the DPO loss $\mathcal{L}_{t}$ for the text semantics preference data $\mathcal{P}_{t}$.

For $\mathcal{P}_{s}$, directly using the same method as in Eq.\ref{eq_dpo} and Eq.\ref{eq_logpi} to compute the DPO loss $\mathcal{L}_{s}$ is challenging, since it is infeasible to calculate the likelihood of the LVLM generating a segmentation embedding $f$. To address this issue, we propose an alternative approach that computes the preference optimization loss by assessing the similarity among different $\{{f}^{n, i}\}_{i=1}^{3}$, which are embeddings corresponding to the $n$-th segmentation target in the LVLM's response derived from different perturbed images $\{\hat{I}^{i}\}_{i=1}^{3}$. Formally,
\begin{equation} \label{eq_seg_dpo}
\begin{aligned}
    &r_{w}^{n} = \beta_{s} \left(\cos(f_{\theta}^{n,L_{s}^{n}[1]}, f_{{\rm ref}}^{n,L_{s}^{n}[0]}) - \cos(f_{{\rm ref}}^{n,L_{s}^{n}[1]}, f_{{\rm ref}}^{n,L_{s}^{n}[0]})\right),\\
    &r_{l}^{n} = \beta_{s} \left(\cos(f_{\theta}^{n,L_{s}^{n}[1]}, f_{{\rm ref}}^{n,L_{s}^{n}[2]})-\cos(f_{{\rm ref}}^{n,L_{s}^{n}[1]}, f_{{\rm ref}}^{n,L_{s}^{n}[2]})\right),\\
    &\mathcal{L}_{s} = -\mathbb{E}_{\mathcal{P}_{s}}\frac{1}{N}\sum_{n=1}^{N}\log \sigma(r_{w}^{n} - r_{l}^{n})\mathds{1}(M_{{\rm ref}}^{n, L_{s}^{n}[0]} \succeq M_{\theta}^{n, L_{s}^{n}[1]}),
\end{aligned}
\end{equation}
where $\beta_{s}$ is a hyperparameter, $\cos$ denotes cosine similarity, $N$ is the number of segmentation targets in a response, $L_{s}^{n}$ is an index list sorted by segmentation performance, for example, $L_{s}^{n} = [3, 1, 2]$ if the respective segmentation masks from reference model $M_{{\rm ref}}^{n, 3}$ surpasses $M_{{\rm ref}}^{n, 1}$ and $M_{{\rm ref}}^{n, 1}$ surpasses $M_{{\rm ref}}^{n, 2}$. $M_{{\rm ref}}^{n, L_{s}^{n}[0]} \succeq M_{\theta}^{n, L_{s}^{n}[1]}$ refers to $M_{{\rm ref}}^{n, L_{s}^{n}[0]}$ outperforming $M_{\theta}^{n, L_{s}^{n}[1]}$, i.e., with a higher $s^{i}$ (Eq.\ref{eq_select_score}) in the first half of finetuning or a higher boundary IoU in the latter half. Through this function, we encourage $f^{n,L_{s}^{n}[1]}$ (with the medium segmentation performance) to move further away from $f^{n,L_{s}^{n}[2]}$ (with the worst performance) and closer to $f^{n,L_{s}^{n}[0]}$ (with the best performance). Note that the loss is set to zero if the segmentation $M_{\theta}^{n, L_{s}^{n}[1]}$ from the finetuned policy model has already suppressed $M_{{\rm ref}}^{n, L_{s}^{n}[0]}$, as continuing to optimize $f_{\theta}^{n,L_{s}^{n}[1]}$ to close the distance to the worse-performing $f_{{\rm ref}}^{n, L_{s}^{n}[0]}$ would be detrimental. 

Finally, the overall preference optimization loss $\mathcal{L}_{pre}$ is computed as $\mathcal{L}_{pre} = \mathcal{L}_{t} + \mathcal{L}_{s}$, which is employed in the training process detailed in Sec.\ref{text_overall_process} for finetuning. 

\begin{figure}[t]
    \centering
    \includegraphics[width=0.9\linewidth]{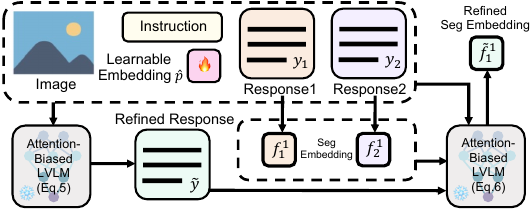}
    \caption{Illustration of \textbf{preference-based ensemble}. For simplify of illustration, in this figure, the number $K$ of the generated responses is 2, the number $N$ of segmentation targets is 1.}
    \vspace{-0.5\baselineskip}
    \label{ensemble_fig}
\end{figure}

\subsection{Preference-Based Ensemble} \label{text_preference_ensemble}
After completing the preference optimization using the above method, we propose a preference-based ensemble mechanism to further improve the reliability of the model’s responses. As shown in Figure \ref{ensemble_fig}, this mechanism integrates multiple outputs from the LVLM and, like our preference-based optimization method, is specially designed to focus on both text semantics and segmentation accuracy. 

Specifically, the LVLM has the capability to generate multiple distinct text responses for a given image-instruction pair $(I, x)$ due to the inherent randomness in its decoding process. Our empirical observations reveal that hallucinations within these different responses typically vary in location. For instance, response $y_{1}$ might exhibit no hallucinations in the first sentence but contain errors in subsequent ones, whereas response $y_{2}$ might display the reverse pattern. Leveraging this observation, we employ the LVLM to integrate various responses for refinement. Specifically, we first use the LVLM to generate $K$ different responses $\{y_{k}\}_{k=1}^{K}$. Optimized with the DPO loss detailed in Eq.\ref{eq_dpo}, the likelihood of tokens in the response can reflect the extent to which they align with human preferences. Leveraging this insight and to consider both local and global characteristics, we calculate a preference score $\tau_{k}^{i}$ for each token $y_{k}^{i}$ in the response $y_{k}$ by summing the likelihood of $y_{k}^{i}$ with the average likelihood of all tokens in the sentence to which $y_{k}^{i}$ belongs, followed by normalization to the range of [-1,1]. We then concatenate the input $(I, x)$, responses $\{y_{k}\}_{k=1}^{K}$, along with a set of learnable prompt embedding $\hat{p}$ and feed them into the LVLM for generating a refined response $\tilde{y}$. During this process, attentions in the LVLM are modified to focus more on tokens with higher $\tau_{k}^{i}$, as they are more likely to contain correct information preferred by humans with fewer hallucinations. Specifically, with the input $E = [e_{I}, e_{x}, \{e_{y_{k}}\}_{k=1}^{K}, e_{\hat{p}}]$, where each item refers to the token embedding of $I$, $x$, $\{y_{k}\}_{k=1}^{K}$ and $\hat{p}$, respectively, each attention matrix $A$ in the LVLM is rewritten as:
\begin{equation} \label{eq_ensemble_attention}
\begin{aligned}
    &A = {\rm Softmax}\left(\textbf{Q}(E)\cdot\textbf{K}(E)^{\rm T} / \sqrt{d_{k}} + \gamma \right),\\
    & \gamma_{j} = \sigma\left(\tau_{k}^{i}\right) - 0.5 \enspace
    \text{if}\enspace E_{j}\; \text{is}\; e_{y_{k}^{i}}\enspace \text{else}\enspace 0,
\end{aligned}
\end{equation}
where $\gamma_{j}$ refers to the $j$-th row on $\gamma$, $\sigma$ denotes the Sigmoid function, $\textbf{Q}(E)$ and $\textbf{K}(E)$ are the query and key features derived from $E$, $d_{k}$ refers to the channel dimension of $E$. 

After obtaining the refined response $\tilde{y}$ from the LVLM's output, the next step is to extract the segmentation embedding $\tilde{f}$ for each \texttt{<seg>} token in $\tilde{y}$. To implement this, we first derive segmentation embeddings $\{\{f_{k}^{n}\}_{n=1}^{N}\}_{k=1}^{K}$ from all responses $\{y_{k}\}_{k=1}^{K}$ ($N$ denotes the number of segmentation targets in each $y$), then employ a similar method used for $\tilde{y}$ to compute $\tilde{f}$ by integrating information from $\{\{f_{k}^{n}\}_{n=1}^{N}\}_{k=1}^{K}$ to enhance segmentation accuracy. Our empirical analysis shown in Supp indicates that a sentence’s preference score $\eta$, represented by the average prediction likelihood of all tokens in it, is positively correlated with the accuracy of the segmentation target contained in that sentence. Inspired by this finding, we calculate such a preference score $\eta_{k}^{n}$ for the sentence to which each $f_{k}^{n}$'s \texttt{<seg>} token belongs followed by normalization to the range [-1, 1] , and then feed the concatenation of $\{I, x, \{y_{k}\}_{k=1}^{K}, \hat{p}, \{\{f_{k}^{n}\}_{n=1}^{N}\}_{k=1}^{K}, \tilde{y}\}$ into the LVLM to generate the refined $\tilde{f}$. In this process, we follow the same method as in Eq.\ref{eq_ensemble_attention} to adjust the attention so that the LVLM focuses more on high-$\eta$ $f_{k}^{n}$ with higher reliability, with $\gamma$ in Eq.\ref{eq_ensemble_attention} rewritten as:
\begin{equation}
    \gamma_{j} = \sigma\left(\eta_{k}^{n}\right)- 0.5\enspace 
    \text{if}\enspace E_{j}\; \text{is}\; e_{f_{k}^{n}}\enspace \text{else}\enspace 0,
\end{equation}
where $e_{f_{k}^{n}}$ denotes the token embedding for $f_{k}^{n}$. Finally, $\tilde{f}$ generated from the LVLM is fed into the segmentation decoder to produce the segmentation mask.  

\subsection{Overall Process of Training and Inference}\label{text_overall_process} After introducing the proposed preference-based optimization and ensemble methods, we then present the overall process for model training and inference in this section. The training process consists of three stages: First, the model is supervised finetuned (SFT) using the same method as PixelLM. Next, the segmentation decoder is frozen, and the LVLM is finetuned using the preference data collected in Sec.\ref{text_optimization}. Note that different types of the segmentation embedding preference $\mathcal{P}_{s}$ are collected and used in the first and second halves of this stage (see Sec.\ref{text_optimization} for details). The loss in this stage is the sum of the preference optimization loss $\mathcal{L}_{pre}$ described in Sec.\ref{text_optimization} and the cross-entropy loss $\mathcal{L}_{ce}$ for segmentation masks, i.e., $\mathcal{L}_{pre}+\mathcal{L}_{ce}$. Finally, the model is finetuned to optimize the preference-based ensemble ability illustrated in Sec.\ref{text_preference_ensemble}, using a loss detailed in Supp that is specifically designed to ensure the improvement of the refined text response and segmentation compared to the original ones. Note that in this stage, only the learnable prompt embedding $\hat{p}$ is updated, while all other parameters, including the LVLM and decoder, are frozen to prevent losing the capabilities gained through preference optimization.

During inference, we employ the ensemble method in Sec.\ref{text_preference_ensemble}, generating $K$ different responses and integrating them for refinement and producing the final result.

\section{Experiments}
\subsection{Experimental Settings}
\noindent \textbf{Implementation Details. }We conduct experiments based on the model architecture of PixelLM \cite{ren2024pixellm}, with the pretrained LLaVA-7B and LLaVA-llama2-13B as the LVLM and the CLIP-ViT-L/14-336 model as the vision encoder. The number $N_{p}$ of the generated perturbed images for segmentation embedding preference is 30, $\beta_{s}$ in Eq.\ref{eq_seg_dpo} is set to 10, and the number $K$ of generated responses in the preference-based ensemble method is 3. In the overall training process described in Sec.\ref{text_overall_process}, the supervised finetuning stage follows the exact same training hyperparameter settings as PixelLM, training on a combination of multiple datasets including ADE20K \cite{zhou2017scene}, COCO-Stuff \cite{caesar2018coco}, LVIS-PACO \cite{ramanathan2023paco}, refCOCO series \cite{kazemzadeh2014referitgame}, LLAVA-150k \cite{liu2023llava}, and MUSE \cite{ren2024pixellm} for 10 epochs. Both the preference-based optimization stage and the stage for optimizing preference-based ensemble ability are carried out on MUSE for 2 epochs. Please see Supp for more details of hyperparameter settings. 

\noindent \textbf{Evaluation Metrics. }We follow the same method as PixelLM by using a GPT-assisted approach for evaluation, which considers both the alignment between the text description and the predicted objects, as well as the accuracy of the segmentation masks. The gIoU and cIoU scores are calculated based on this evaluation method. Readers can refer to \cite{ren2024pixellm} for more details. Additionally, we use two other methods to evaluate the quality and degree of hallucination in the generated text responses. First, the CHAIR metric \cite{rohrbach2018object} is employed to assess the proportion of objects present in the generated response but absent in the ground truth, which contains two sub-metrics $C_{S}$ and $C_{I}$ computed as:
\begin{equation}
\scriptsize
    C_{S} = \frac{|\{\text{\scriptsize responses w/ hallucinated objects}\}|}{|\{\text{\scriptsize all responses}\}|}, C_{I} = \frac{|\{\text{\scriptsize hallucinated objects}\}|}{|\{\text{\scriptsize all mentioned objects}\}|}.
\end{equation}
The CHAIR metric can only reflect the degree of object hallucination. For a more comprehensive evaluation, we also employ the method from \cite{huang2024opera}, where we prompt ChatGPT to evaluate the correctness of the LVLM’s response given the input image-instruction pair. In this way, a score is generated from ChatGPT to assess the quality of the response. Please see Supp for the detailed prompt used in this method.

\begin{table*}[t]\centering
    \renewcommand{\arraystretch}{0.95}
    \begin{adjustbox}{width=1\linewidth,center}
    \setlength\tabcolsep{5pt}

        \begin{tabular}{c|l|ccccc|cc|cc|ccccc}
            \toprule
            
            \multirow{3}{*}{\makecell{LLM\\Size}} & \multirow{3}{*}{Method} & \multicolumn{5}{c|}{Val} & \multicolumn{9}{c}{Test}   \\

            \cline{3-16}
            & ~  & \multicolumn{5}{c|}{overall}  & \multicolumn{2}{c|}{few targets} & \multicolumn{2}{c|}{many targets} & \multicolumn{5}{c}{overall}  \\ 

            \cline{3-16}
            
            & ~  & gIoU $\uparrow$ & cIoU $\uparrow$ & $C_{S}$ $\downarrow$ & $C_{I}$ $\downarrow$ & Score $\uparrow$ & gIoU $\uparrow$& cIoU $\uparrow$ &gIoU $\uparrow$ &cIoU $\uparrow$ &gIoU $\uparrow$ &cIoU $\uparrow$ & $C_{S}$ $\downarrow$ & $C_{I}$ $\downarrow$ & Score $\uparrow$   \\ 
            
            \hline
            
            \multirow{6}{*}{7B} & LISA \cite{lai2024lisa} & 17.2 &28.8 &23.2 & 10.3 & 5.6  &24.4 &36.5  &9.6 &24.5  &12.8  &27.1 & 24.1 & 10.8 & 5.2   \\

             & GSVA \cite{xia2024gsva} &38.9 &40.9 &21.8 & 9.9 & 6.3 & 44.3  &54.1 &34.1 &38.2  &36.3 &41.6 &22.7 & 10.1 & 6.0 \\

              & GLaMM \cite{rasheed2024glamm} & 41.5 & 48.0 & 20.8 & 9.5 & 6.1 & 44.4 & 57.9 & 36.4 & 40.9 & 38.1 & 44.5 & 24.7 & 9.6 & 6.0 \\

            & PixelLM \cite{ren2024pixellm} &41.9  &48.9 &22.0 & 9.8 & 6.2  &44.0 &57.8  &37.3 &42.3  &38.7 &45.6 &22.2 & 9.6 & 6.2 \\
            \cline{2-16}
             & \cellcolor{mygray}POPEN$\dagger$ & \cellcolor{mygray}44.1& \cellcolor{mygray}53.8 & \cellcolor{mygray}12.1 & \cellcolor{mygray}5.6 & \cellcolor{mygray}7.2 & \cellcolor{mygray}45.7 & \cellcolor{mygray}61.6 & \cellcolor{mygray}40.2 & \cellcolor{mygray}46.7 & \cellcolor{mygray}41.3 & \cellcolor{mygray}49.9 & \cellcolor{mygray}12.2 & \cellcolor{mygray}5.8 & \cellcolor{mygray}7.0\\
            & \cellcolor{mygray}POPEN & \cellcolor{mygray}\textbf{45.4} & \cellcolor{mygray}\textbf{55.2} & \cellcolor{mygray}\textbf{9.3} & \cellcolor{mygray}\textbf{4.3} & \cellcolor{mygray}\textbf{7.7} & \cellcolor{mygray}\textbf{46.4} & \cellcolor{mygray}\textbf{62.9} & \cellcolor{mygray}\textbf{41.3} & \cellcolor{mygray}\textbf{48.1} & \cellcolor{mygray}\textbf{42.4} & \cellcolor{mygray}\textbf{51.2} & \cellcolor{mygray}\textbf{9.5} & \cellcolor{mygray}\textbf{4.3} & \cellcolor{mygray}\textbf{7.4}\\

            \cline{1-16}
            
            \multirow{5}{*}{13B} & LISA \cite{lai2024lisa}
            &20.0  &28.9 &22.0 & 10.5 & 5.9  &27.3  &38.2 & 10.7  &25.6 &14.2 &28.3 &23.5 & 10.2 & 5.3  \\
            & GSVA \cite{xia2024gsva} & 41.7 & 50.3 &20.6 & 9.3 & 6.4 & 45.1 & 62.5 & 39.5 & 45.1 & 40.7 & 48.8 &21.1 & 9.5 & 6.4 \\

            & PixelLM \cite{ren2024pixellm} &44.0  &52.9 &21.1 & 9.4 & 6.4  &45.0  &61.9  &41.6  &47.9  &42.3 &50.9 &22.0 & 9.6 & 6.6  \\
            \cline{2-16}
             & \cellcolor{mygray}POPEN$\dagger$ & \cellcolor{mygray}46.9 & \cellcolor{mygray}57.7 & \cellcolor{mygray}11.9 & \cellcolor{mygray}5.6 & \cellcolor{mygray}7.3 & \cellcolor{mygray}47.4 & \cellcolor{mygray}66.6 & \cellcolor{mygray}44.1 & \cellcolor{mygray}52.2 & \cellcolor{mygray}44.7 & \cellcolor{mygray}55.3 & \cellcolor{mygray}12.5 & \cellcolor{mygray}5.6 & \cellcolor{mygray}7.4\\
            & \cellcolor{mygray}POPEN & \cellcolor{mygray}\textbf{48.0} & \cellcolor{mygray}\textbf{59.1} & \cellcolor{mygray}\textbf{9.1} & \cellcolor{mygray}\textbf{4.2} & \cellcolor{mygray}\textbf{7.7} & \cellcolor{mygray}\textbf{48.3} & \cellcolor{mygray}\textbf{67.9} & \cellcolor{mygray}\textbf{45.5} & \cellcolor{mygray}\textbf{53.9} & \cellcolor{mygray}\textbf{46.0} & \cellcolor{mygray}\textbf{56.9} & \cellcolor{mygray}\textbf{9.1} & \cellcolor{mygray}\textbf{4.4} & \cellcolor{mygray}\textbf{7.9}\\
            \bottomrule            
        \end{tabular}
        \end{adjustbox}
        \vspace{-0.5\baselineskip}
    \caption{Comparison on MUSE benchmark. POPEN$\dagger$ refers to our method w/o preference-based ensemble. Score refers to the evaluation scores from ChatGPT. Note that the results for LISA and PixelLM are reproduced by us and differ from those reported in \cite{ren2024pixellm}, which may be due to the use of different ChatGPT versions for calculating gIoU and cIoU.
      }
      \vspace{-0.5\baselineskip}
    \label{tabel_comp_muse}   
\end{table*}

\subsection{Main Results}
\noindent \textbf{Comparison on MUSE. }We compare our approach with other methods on the reasoning segmentation task. Results for both segmentation-related metrics, including gIoU and cIoU, as well as text-related metrics including $C_{S}$, $S_{I}$ and GPT-score, are presented in Table \ref{tabel_comp_muse}. Among the compared methods, LISA \cite{lai2024lisa} is the pioneering approach in this field but has relatively poor performance, primarily due to its limitation to segment only one single target object per input. GSVA \cite{xia2024gsva} addresses this issue by introducing multiple segmentation tokens and thus being able to handle multiple target objects at once. PixelLM \cite{ren2024pixellm} further improves the performance by employing a better segmentation feature extraction method and a stronger decoder. Benefiting from the task-tailored and innovatively proposed preference optimization method in this work, our method achieves significant improvements in both the quality of the text response and the accuracy of the segmentation mask compared to LISA, GSVA, GLaMM and PixelLM. Furthermore, after applying the proposed preference-based ensemble method to fuse multiple outputs, our performance advantage becomes even more pronounced. These results demonstrate the significant superiority of our method compared to previous SOTA LVLM-based segmentation methods.

\begin{table}[t]
\renewcommand{\arraystretch}{0.95}
    \centering
    \resizebox{\columnwidth}{!}
    {
    \setlength\tabcolsep{4pt}
        \begin{tabular}{l|ccc|ccc|cc}
            \toprule
            
            \multirow{2}*{Method} & \multicolumn{3}{c|}{refCOCO} & \multicolumn{3}{c|}{refCOCO+}  & \multicolumn{2}{c}{refCOCOg}  \\ 
            
            \cline{2-9}
            
            ~  & val & testA & testB & val & testA & testB & val(U) & test(U)  \\ 
        
            \hline
            
            MCN~\citep{luo2020multi} & 62.4 & 64.2 & 59.7 & 50.6 & 55.0 & 44.7 & 49.2 & 49.4 \\

            VLT~\citep{ding2021vision} & 67.5 & 70.5 & 65.2 & 56.3 & 61.0 & 50.1 & 55.0 & 57.7 \\

            CRIS~\citep{wang2022cris} &70.5 & 73.2 & 66.1 & 62.3 & 68.1 & 53.7 & 59.9 & 60.4 \\

            LAVT~\citep{yang2022lavt} & 72.7 & 75.8 & 68.8 & 62.1 & 68.4 & 55.1 & 61.2 & 62.1 \\
            
            ReLA~\citep{liu2023gres} & 73.8 & 76.5 & 70.2 & 66.0 & 71.0 & 57.7 & 65.0 & 66.0 \\
            
            X-Decoder~\citep{zou2023generalized} & - & - & - & - & - & - & 64.6 & -  \\

            SEEM~\citep{zou2024segment} & - & - & - & - & - & - & 65.7 & -    \\
            
            \hline
            
            LISA~\cite{lai2024lisa}  & 74.1 & 76.5 & 71.1 & 62.4 & 67.4 & 56.5 & 66.4 & 68.5 
            \\

            PixelLM \cite{ren2024pixellm} &73.0   &76.5   &68.2  & 66.3 & 71.7    &58.3 &69.3   & 70.5   \\
            GSVA \cite{xia2024gsva} & 76.4 & 77.4 & 72.8 & 64.5 &67.7 & 58.6 &71.1 &72.0\\
            \rowcolor{mygray} POPEN & \textbf{78.5} & \textbf{79.9} & \textbf{73.0} & \textbf{70.3} & \textbf{74.4} & \textbf{62.4} & \textbf{73.8} & \textbf{74.6}\\
            \midrule
            LISA (ft) \cite{lai2024lisa} & 74.9 & 79.1 & 72.3 & 65.1 & 70.8 & 58.1 & 67.9 & 70.6\\
            GSVA (ft) \cite{xia2024gsva} & 77.2 & 78.9 & 73.5 & 65.9 & 69.6 & 59.8 & 72.7 & 73.3\\
            \rowcolor{mygray} POPEN (ft) & \textbf{79.3} & \textbf{82.0} & \textbf{74.1} & \textbf{73.1} & \textbf{77.0} & \textbf{65.1} & \textbf{75.4} & \textbf{75.6}\\
            \bottomrule 
        \end{tabular}
    }
    \vspace{-0.5\baselineskip}
    \caption{Results on referring expression segmentation. `ft' refers to finetuning on referring expression segmentation datasets.} 
    \vspace{-0.5\baselineskip}
    \label{table:refer_seg}   
\end{table}

\noindent \textbf{Comparison on Referring Expression Segmentation.}We further evaluate our method on datasets for referring expression segmentation (RES), including refCOCO and the more challenging refCOCO+ and refCOCOg. Compared to both the traditional RES methods and LVLM-based methods like LISA and PixelLM, our approach achieves the best performance on all datasets. These results demonstrate the high effectiveness of our method for RES.

In \textbf{Supp}, we present results on more benchmarks like the grounded conversation generation task on \textbf{GranD$_f$} \cite{rasheed2024glamm}.

\subsection{Ablation Study} \label{text_ablation}
In this section, we conduct experiments on MUSE validation set to evaluate the effectiveness of our designs in this work. Both the segmentation metrics (gIoU, cIoU) and text metrics ($C_{S}$, $C_{I}$) are reported. Due to paper length limitation, more ablation study results, including the evaluation for hyperparameters in our method, are presented in \textbf{Supp}.

\begin{table}[t]
    \centering
    \setlength\tabcolsep{4pt}
    \resizebox{1\linewidth}{!}{
    \begin{tabular}{l|cccc}
        \toprule
        Method & gIoU $\uparrow$ & cIoU $\uparrow$ & $C_{S}$ $\downarrow$ & $C_{I}$ $\downarrow$\\
        \midrule
         POPEN & \textbf{45.42} & \textbf{55.20} & \textbf{9.29} & \textbf{4.31}\\
         \midrule
         POPEN w/o preference-based optimization & 42.47 & 49.83 & 20.15 & 9.29\\
         POPEN w/o preference-based ensemble & 44.10 & 53.76 & 12.09 & 5.62\\
         \bottomrule
    \end{tabular}}
    \vspace{-0.5\baselineskip}
     \caption{Ablation study of two main components in POPEN.} 
    \label{ablation_component}
\end{table}

\noindent \textbf{Effectiveness of Different Components. }We first evaluate two main components of our proposed POPEN: preference-based optimization and preference-based ensemble. As shown in Table \ref{ablation_component}, removing either of these components can lead to a significant decrease in both the quality of the text response and the segmentation accuracy, demonstrating their high effectiveness and importance.

\begin{table}[t]
    \centering
    \setlength\tabcolsep{4pt}
    \resizebox{1\linewidth}{!}{
    \begin{tabular}{l|cccc}
        \toprule
        Method & gIoU $\uparrow$ & cIoU $\uparrow$ & $C_{S}$ $\downarrow$ & $C_{I}$ $\downarrow$\\
        \midrule
         POPEN & \textbf{45.42} & \textbf{55.20} & \textbf{9.29} & \textbf{4.31}\\
         \midrule
         POPEN w/o $\mathcal{L}_{t}$ (Eq.\ref{eq_dpo})& 44.62 & 54.17 & 19.75 & 9.08\\
         POPEN w/o $\mathcal{L}_{s}$ (Eq.\ref{eq_seg_dpo}) & 42.80 & 50.39 & 10.41 & 4.95\\
         \midrule
         POPEN w/o $\lambda$ in Eq.\ref{eq_logpi} & 45.02 & 54.62 & 13.56 & 6.09\\
         POEPN w/o $\mathds{1}(M_{{\rm ref}}^{n, L_{s}^{n}[0]} \succeq M_{\theta}^{n, L_{s}^{n}[1]})$ in Eq.\ref{eq_seg_dpo} & 43.97 & 53.89 & 9.90 & 4.68\\
         \bottomrule
    \end{tabular}}
     \caption{Ablation study of preference-based optimization.}
     \vspace{-0.5\baselineskip}
    \label{table_ablation_optim}
\end{table}

\begin{figure*}[t]
    \centering
    \includegraphics[width=1\linewidth]{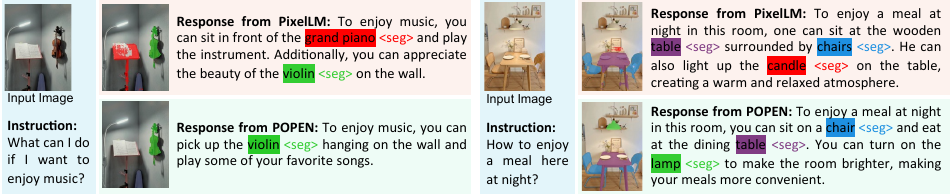}
    \caption{Comparative examples of text responses and segmentation results between PixelLM and our POPEN.}
    \vspace{-0.5\baselineskip}
    \label{example_fig}
\end{figure*}

\noindent \textbf{Ablation Study of Preference-based Optimization.} In our method, the loss used in the preference-based optimization (Sec.\ref{text_optimization}) is the sum of two functions: the text DPO loss $\mathcal{L}_{t}$ (Eq.\ref{eq_dpo}) and the segmentation DPO loss $\mathcal{L}_{s}$ (Eq.\ref{eq_seg_dpo}). As shown in Table \ref{table_ablation_optim}, removing either of these losses significantly reduces the model's performance. Notably, excluding the text-based $\mathcal{L}_{t}$ or the segmentation-based $\mathcal{L}_{s}$ affects not only the corresponding text quality or segmentation accuracy but both. This could be because the LVLM can mutually benefit from learning from both the segmentation and text generation tasks. Specifically, better text generation reduces hallucinations, preventing segmentation of incorrect objects; stronger segmentation ability enhances the LVLM’s target localization capability, which in turn prevents the generation of hallucinatory text that includes objects beyond the target. Therefore, the combined use of $\mathcal{L}_{t}$ and $\mathcal{L}_{s}$ is crucial for reasoning segmentation, which requires the simultaneous accuracy of both text and segmentation. In addition, we also validate the design details of $\mathcal{L}_{t}$ and $\mathcal{L}_{s}$, including (1) $\lambda$ in Eq.\ref{eq_logpi} and (2) $\mathds{1}(M_{{\rm ref}}^{n, L_{s}^{n}[0]} \succeq M_{\theta}^{n, L_{s}^{n}[1]})$ in Eq.\ref{eq_seg_dpo}. Excluding these elements leads to a decrease in performance, demonstrating the effectiveness of our designs.

\begin{table}[t]
    \centering
    \setlength\tabcolsep{5pt}
    \resizebox{1\linewidth}{!}{
    \begin{tabular}{l|cccc}
        \toprule
        Collection Method & gIoU $\uparrow$ & cIoU $\uparrow$ & $C_{S}$ $\downarrow$ & $C_{I}$ $\downarrow$\\
        \midrule
        Curriculum Collection & \textbf{45.42} & \textbf{55.20} & \textbf{9.29} & \textbf{4.31}\\
        \midrule
         Random & 43.06 & 51.35 & 9.78 & 4.62\\
         Based on $s$ (Eq.\ref{eq_select_score}) for both halves & 44.10 & 53.73 & 9.51 & 4.45\\
         Based on boundary IoU for both halves & 43.11 & 51.30 & 9.90 & 4.67\\
         \bottomrule
    \end{tabular}}
     \caption{Effectiveness of different methods for segmentation preference data collection. ``Based on $s$" and ``Based on boundary IoU" respectively refer to the collection methods for the 1st and 2nd halves of finetuning used in our POPEN (details in Sec.\ref{text_optim}).}
     \vspace{-0.5\baselineskip}
    \label{table_ablation_collection}
\end{table}

\noindent \textbf{Effectiveness of Curriculum Collection for $\mathcal{P}_{s}$.} As detailed in Sec.\ref{text_optim}, we employ a curriculum method for obtaining segmentation embedding preference data $\mathcal{P}_{s}$, collecting different types of $\mathcal{P}_{s}$ for the first and second halves of finetuning. As shown in Table \ref{table_ablation_collection}, when this curriculum method is replaced by the random collection, or when the same strategy is used for data collection in both halves, the model’s performance significantly decreases, demonstrating the high effectiveness of our proposed method.

\noindent \textbf{Ablation Study of Preference-based Ensemble. }We further conduct ablation study to evaluate the following components of our proposed preference-based ensemble method (Sec.\ref{text_preference_ensemble}): (1) the text response ensemble, (2) the segmentation embedding ensemble, (3) $\gamma$ in Eq.\ref{eq_ensemble_attention} to focus on high-reliability components, and (4) an additional learnable prompt embedding $\hat{p}$ for LVLM's input. In addition, we also validate the two elements that are summed to form $\tau_{k}^{i}$ in Eq.\ref{eq_ensemble_attention} -- the likelihood $p_{k}^{i}$ of token $y_{k}^{i}$ and the average likelihood of all tokens in the sentence to which $y_{k}^{i}$ belongs. The results presented in Table \ref{table_ablation_ensemble} indicate that all these components and designs can contribute significantly to the performance improvement, demonstrating the soundness and effectiveness of our method.

\begin{table}[t]
    \centering
    \setlength\tabcolsep{1.8pt}
    \resizebox{1\linewidth}{!}{
    \begin{tabular}{l|cccc}
        \toprule
        Collection Method & gIoU $\uparrow$ & cIoU $\uparrow$ & $C_{S}$ $\downarrow$ & $C_{I}$ $\downarrow$\\
        \midrule
        POPEN & \textbf{45.42} & \textbf{55.20} & \textbf{9.29} & \textbf{4.31}\\
        \midrule
        POPEN w/o text response ensemble & 44.90 & 54.62 & 11.89 & 5.50\\
         POPEN w/o segmentation embedding ensemble & 44.33 & 54.07 & 9.55 & 4.41\\
          POPEN w/o $\gamma$ in Eq.\ref{eq_ensemble_attention} & 44.71 & 54.35 & 11.07 & 5.11\\
          POPEN w/o additional learnable embedding $\hat{p}$ & 43.89 & 53.67 & 13.01 & 5.75\\
          \midrule
          $\tau_{k}^{i}$ w/o likelihood $p_{k}^{i}$ of token $y_{k}^{i}$ & 45.09 & 54.77 & 9.80 & 4.54\\
          $\tau_{k}^{i}$ w/o average $p$ of all tokens in $y_{k}^{i}$'s sentence& 45.00 & 54.65 & 11.06 & 5.23\\
         \bottomrule
    \end{tabular}}
     \caption{Ablation study of preference-based ensemble.}
     \vspace{-0.5\baselineskip}
    \label{table_ablation_ensemble}
\end{table}

\vspace{-0.5\baselineskip}
\subsection{Qualitative Comparison}
\vspace{-0.5\baselineskip}
In Figure \ref{example_fig}, we present examples comparing text responses and segmentation results between our POPEN and PixelLM \cite{ren2024pixellm}. In these examples, PixelLM suffers from serious hallucinations, generating objects in its text responses that do not exist within the images, such as the ``grand piano" in the left example and ``candle" in the right example. Furthermore, the segmentation accuracy is suboptimal, with coarse details for the segmentation of ``table" and ``chair" in the right example (failing to segment the table's left leg). By employing the proposed preference-based optimization and ensemble methods, our POPEN achieves significantly improved results, effectively mitigating hallucination in text responses and enhancing segmentation accuracy. These comparative results demonstrate the high effectiveness and advantage of our method compared to PixelLM.

\vspace{-0.5\baselineskip}
\section{Conclusion}
\vspace{-0.5\baselineskip}
This paper proposes POPEN, a new framework that incorporates innovatively proposed preference-based optimization and ensemble methods with task-tailored designs, significantly improving the LVLM’s ability to handle reasoning segmentation. Extensive experiments demonstrate the effectiveness of our proposed method. We consider our POPEN an important step toward aligning the pixel-level understanding capabilities in LVLMs with human preferences for performance improvement.\\

\noindent \textbf{Acknowledgement}
T. C. acknowledges support from ZJU Kunpeng \& Ascend Center of Excellence.

\clearpage
{\small
\bibliographystyle{unsrt}
\bibliography{egbib}
}

\appendix

\vspace{5\baselineskip}
\section{Discussion of Computation}
\noindent \textbf{Training Time. }As detailed in Sec.3.4 of the  main paper, our method consists of three training stages. The first stage follows the method of PixelLM \cite{ren2024pixellm}, training for 10 epochs, which takes approximately 1.5 days for the 7B model on 8 A100 GPUs. The second and third stages train for 2 epochs each, requiring approximately 5 hours and 8 hours, respectively.\\

\noindent \textbf{Inference Time. }Our proposed preference-based ensemble method needs to generate $K$ different responses and fuses them. In our experiments, $K$ is set to 3. Theoretically, this would require 4 times the computation compared to the the original PixelLM w/o ensemble. However, benefiting from optimizations such as parallel computation and KV cache, in practice, the averge inference time of our method is only 1.57 times that of the method w/o ensemble. This is entirely acceptable considering the significant improvement brought by our preference-based ensemble approach. Also note that even without using the ensemble method that requires additional computational cost, our method can still significantly outperform the baseline PixelLM, as shown by the results for POPEN$\dagger$ in Table.1 of main paper. This further demonstrates the superiority of our approach.

\section{More Details of Proposed Method}

\noindent \textbf{ChatGPT Prompt for Response Correction.}
As introduced in Sec.3.2 of main paper, we use ChatGPT to refine LVLM's response $y$ by modifying, adding, or deleting certain words or sentences in $y$, thus generating a corrected response $y_{c}$ with fewer errors to construct the text semantic preference data. The ChatGPT prompt format for this operation is as follows: 

\textit{You are an assistant designed to help me correct an incorrect answer to a question about an image. I will provide you with an image, a question, an answer from an LVLM, and an object list. You need to modify, add, or delete certain words or sentences in the LVLM’s answer to correct mistakes, including incorrect objects and faulty reasoning. The corrected answer should include only the objects in the object list. You should return: (1) The original LVLM's answer I provided, in which you should mark the deleted or modified parts in the answer in quotes. (2) Your corrected answer, in which you should mark the modified or added parts compared to the original answer in quotes. Please ensure that only the modified, deleted, or added parts are marked. Do not mark synonyms as modifications. Please retain the sentence structure and content of the LVLM’s original answer as much as possible, without adding extra information beyond what is necessary for correction.}

In this prompt, the object list refers to a list containing the names of all objects within the ground truth response. \\

\noindent \textbf{ChatGPT Prompt to Intentionally Introduce Errors.}As introduced in Sec.3.2 of the main paper, to enrich dataset, for some of the LVLM’s responses that contain only few errors, we instruct ChatGPT to intentionally introduce errors into the ground truth response $y_{g}$ to formulate $y$. Specifically, if ChatGPT finds that an LVLM response has no errors, we use the randomness in decoding to generate three different responses and select one containing errors as $y$ for the text semantics preference. If these responses still contain no errors, we use the following prompt to intentionally introduce errors into the ground truth response $y_{g}$:

\textit{You are an assistant designed to help me intentionally introduce errors into a correct answer to a question about an image. I will provide you with an image, a question, and a correct answer. You need to modify, add, or delete certain words or sentences in the correct answer to introduce some mistakes, such as incorrect objects and faulty reasoning. You should return the modified answer. Please introduce errors into only a small portion of the content (e.g., one or two objects). Please do not perform synonym replacement.}\\

\noindent \textbf{Loss for Preference-Based Ensemble.} As indicated in Sec.3.4 of the main paper, in the third training stage of our method, which aims to optimize the preference-based ensemble capability, we employ a specially designed loss function to ensure that the refined text responses and segmentation outperform the originals. The loss function for this stage is the sum of two components: the text improvement loss $\mathcal{L}_{ti}$ and the segmentation improvement loss $\mathcal{L}_{si}$. To be specific, denote the $K$ generated responses as $\{y_{k}\}_{k=1}^{K}$, the refined response as $\Tilde{y}$ and the ground truth response as $y^{g}$, $\mathcal{L}_{ti}$ is formulated as follows: 
\begin{equation}
\begin{aligned}
    &h_{k} = {\rm BERT}\left(y_{k}\right),\; \Tilde{h} = {\rm BERT}\left(\Tilde{y}\right),\; h^{g} = {\rm BERT}\left(h^{g}\right),\\
    &\mathcal{L}_{ti} = -\mathbb{E}\frac{1}{K}\sum_{k=1}^{K}\log \sigma \left(10 p(\Tilde{y})\left({\rm Cos}(\Tilde{h}, h^{g}) - {\rm Cos}(h_{k}, h^{g})\right)\right),
\end{aligned}
\end{equation}
where $h = {\rm BERT}\left(y\right)$ refers to a feature extracted from BERT with $y$ as input, ${\rm Cos}$ denotes cosine similarity. $p(\Tilde{y})$ refers to the probability of the LVLM generating $\Tilde{y}$. This loss constrains the similarity between the refined response $\Tilde{y}$ and the ground truth response $y^{g}$ to be higher than that between the original responses $y_{k}$ and $y^{g}$, thus optimizing the model to produce more refined response outperforming the original ones. Similarity, the segmentation improvement loss $\mathcal{L}_{si}$ is computed as:
\begin{equation}
\begin{aligned}
    \mathcal{L}_{ti} = -\mathbb{E}\frac{1}{K}\frac{1}{N}\sum_{k=1}^{K}\sum_{n=1}^{N}\log \sigma(10(&{\rm IoU}(\Tilde{M}^{n}, M_{g}^{n})\\-&{\rm IoU}({M}_{k}^{n}, M_{g}^{n}))),
\end{aligned}
\end{equation}
where $N$ is the number of segmentation targets in the response, $\Tilde{M}^{n}$ is the $n$-th refined segmentation mask, $M_{k}^{n}$ is the $n$-th segmentation mask from the $k$-th original response, $M_{g}^{n}$ is the corresponding ground truth mask. 
\\

\begin{table}[t]
    \footnotesize
    \centering
    {
\begin{tabular}{p{3.4cm}|p{2.2cm}}
    \toprule
     config & value\\
    \midrule
    optimizer & AdamW\\
     base learning rate & 3.0e-4\\
     weight decay & 0\\
     optimizer monmentum & $\beta_1$, $\beta_2$=0.9,0.95 \\
     batch size & 16 \\
     learning rate schedule & WarmipDecayLR \\
     warmup iterations & 100 \\
     augmentations & None \\
     \bottomrule
\end{tabular}
}
\caption{Training settings}
\label{table_training_setting}
\vspace{-1\baselineskip}
\end{table}

\noindent \textbf{More Implementation Details.} Some implementation details of our method have been presented in Sec.4.1 of the main paper. Most of the other training settings follow PixelLM and are presented in Table \ref{table_training_setting}. Note that we use the exact same settings shown in Table \ref{table_training_setting} for all three training stages (detailed in main paper Sec.3.4) in our method. The number of learnable prompt embeddings $\hat{p}$ used in the preference-based ensemble is 10.\\

\noindent \textbf{ChatGPT Prompt for Response Evaluation.} As indicated in Sec.4.1 of main paper, for a more comprehensive evaluation of the LVLM's text responses, we prompt ChatGPT to evaluate the correctness of the LVLM’s response given the input image-instruction pair. In this way, a score is generated from ChatGPT to assess the quality of the response. The prompt for ChatGPT in this operation is as follows: 

\textit{I will give you an image, a question and a text response. You are required to score the performance of the text response given the image and question. You should pay extra attention to the hallucination, which refers to the part of responses that are inconsistent with the image content, such as claiming the existence of something not present in the image or describing incorrectly in terms of the counts, positions, or colors of objects in the image. Please rate the responses of the assistants on a scale of 1 to 10, where a higher score indicates better performance, according to the following criteria: (1) Accuracy: whether the response is accurate with respect to the image content and reasoning logic. Responses with fewer hallucinations should be given higher scores. (2) Detailedness: whether the response is rich and complete in necessary details.  Please output a score for such a evaluation. Following the score, please provide an explanation of your evaluation.}

\section{Further Analysis}
\noindent \textbf{Correlation Between Preference Score and Response Quality.} In the proposed preference-based ensemble method, we calculate a preference score based on prediction likelihood to modify each attention matrix in LVLMs, enabling the model to focus more on high-reliability content when integrating multiple text responses. This is based on the property that, after finetuning using the preference optimization method, the prediction likelihood of tokens in the response can reflect the extent to which they align with human preferences. We evaluate this property using the Pearson correlation coefficient. Specifically, for each token $y^{i}$ in the LVLM’s text response $y$, we calculate a score $\tau^{i}$ by summing the likelihood of $y^{i}$ with the average likelihood of all tokens in the sentence to which $y^{i}$ belongs. We then employ ChatGPT to score the preference for each token in the text response based on accuracy, obtaining $c^{i}$ for $y^{i}$. The Pearson correlation coefficient $r$ is then calculated as:
\begin{equation}
r = \frac{\sum_{i=1}^{N_{i}} (\tau_{i} - \bar{\tau})(c_{i} - \bar{c})}{\sqrt{\sum_{i=1}^{N_{i}} (\tau_{i} - \bar{\tau})^2} \sqrt{\sum_{i=1}^{N_{i}} (c_{i} - \bar{c})^2}},
\end{equation}
where $N_{i}$ is the number of tokens in the text response $y$. A higher $r$ indicates a stronger positive correlation between $\tau^{i}$ and the accuracy of the token $y^{i}$. We calculate $r$ across the responses from all image-instruction pairs in the MUSE validation set, and the high average value of 0.76 for $r$ demonstrates the strong correlation. This result highlights the validity of our attention design in Eq.5 of main paper.\\

\begin{SCtable*}[][t]
    \tabstyle{5pt}
    \scalebox{1}{
    
    \begin{tabular}{lccccc|ccccc}
    \toprule
    \multirow{2}{*}{Model} & \multicolumn{5}{c|}{Validation Set} &  \multicolumn{5}{c}{Test Set} \\ 
                           & M & C & AP50 & mIoU & Recall & M & C & AP50 & mIoU & Recall\\ \midrule
    BuboGPT~\cite{zhao2023bubogpt}& 17.2   & 3.6  & 19.1 & 54.0 & 29.4 & 17.1  & 3.5  & 17.3 & 54.1 & 27.0 \\
    Kosmos-2~\cite{peng2023kosmos}& 16.1   & 27.6  & 17.1 & 55.6 & 28.3 & 15.8   & 27.2  & 17.2 & 56.8 & 29.0 \\
    LISA~\cite{lai2024lisa}& 13.0   & 33.9  & 25.2 & 62.0 & 36.3 & 12.9   & 32.2  & 24.8 & 61.7 & 35.5 \\
    
    GLaMM \cite{rasheed2024glamm}               & 16.2   & 47.2  & 30.8 & 66.3 & 41.8 & 15.8 & 43.5  & 29.2 & 65.6 & 40.8 \\ 
    \midrule
    \rowcolor{mygray} POPEN & \textbf{20.3} & \textbf{52.8} & \textbf{34.9} & \textbf{70.1} & \textbf{45.2} & \textbf{20.1} & \textbf{49.4} & \textbf{33.8} & \textbf{69.7} & \textbf{44.0}\\
    \bottomrule
    \end{tabular}}
\caption{\textbf{Performance on the grounded conversation generation (GCG) task of GranD$_f$ Dataset.} Metrics include METEOR (M), CIDEr (C), AP50, mIoU, and Mask Recall. Our POPEN achieves the best performance.}
\label{table_gcg}
\end{SCtable*}

\noindent \textbf{Correlation Between Preference Score and Segmentation Performance.} We use the same method as in the previous section, employing the Pearson correlation coefficient $r$ to measure the correlation between the average prediction likelihood of all tokens in a sentence and the accuracy of the segmentation target contained in the sentence. Across the entire MUSE validation set, the model finetuned with preference optimization achieves a high average  $r$ value of 0.69, indicating a strong positive correlation and demonstrating the rationale behind our designs in Eq. 6 of the main paper for multi-segmentation integration. One possible explanation for this property is that the sequential prediction process of the LVLM would propagate errors and uncertainties from earlier tokens in the text response to subsequent segmentation tokens, while also transmitting errors in the segmentation embedding to later tokens. Consequently, the segmentation accuracy becomes strongly positively correlated with the likelihood of the sentence it belongs to, which reflects the sentence's accuracy and quality.

\section{More Experiments}

\subsection{Comparison on More Benchmarks}

\noindent \textbf{Results on Grounded Conversation Generation of GranD$_f$ Benchmark.} Grounded conversation generation is a task aimed at generating text captions for images as well as segmentation masks for each object within them. To evaluate our method on this task, we follow GLaMM \cite{rasheed2024glamm} by first pretraining the model using the approach described in the main paper, and then finetune it on the GranD$_f$ dataset. The results of the finetuned model on the validation set and test set of GranD$_f$ are presented in Table \ref{table_gcg}. Our POPEN significantly outperforms previous state-of-the-art approaches such as LISA and GLaMM, demonstrating the high effectiveness and superiority of our method.\\

\begin{table}[t]
    \centering
    \setlength\tabcolsep{12pt}
    \resizebox{0.6\linewidth}{!}{
    \begin{tabular}{l|cc}
        \toprule
        Method & gIoU & cIoU \\
        \midrule
        LISA \cite{lai2024lisa} & 52.9 & 54.0\\
        GSVA \cite{xia2024gsva} & 50.5 & 56.4\\
        \midrule
         \rowcolor{mygray} POPEN & \textbf{60.2} & \textbf{64.5}\\
         \bottomrule
    \end{tabular}}
     \caption{Performance on ReasonSeg benchmark.}
    \label{table_reason}
\end{table}

\noindent \textbf{Results on ReasonSeg Benchmark. }We further evaluate our method on the ReasonSeg \cite{lai2024lisa} validation set and compare its performance with LISA and GSVA. The results are presented in Table \ref{table_reason}. Our POPEN achieves the best performance, with significant advantages over the second-best method, showing a +7.3\% improvement on the gIoU metric and +8.1\% on the cIoU metric. These results demonstrate the outstanding performance of our method.\\

\noindent \textbf{Results on Hallucination Benchmarks.} We further evaluate the effectiveness of our method in mitigating hallucination on two hallucination benchmarks, ObjHal and MMHal. As shown in Table \ref{table_hal_bench}, our POPEN outperforms both the baseline LLaVA \cite{liu2023llava} and RLHF-V \cite{yu2024rlhf}, which is specifically designed to address hallucination. This demonstrates the superior effectiveness of our method in mitigating hallucination through the use of additional segmentation training data and the novel techniques we propose in this work.

\begin{table}[t]
    \centering
    \setlength\tabcolsep{8pt}
    \resizebox{0.87\linewidth}{!}{
    \begin{tabular}{l c cc cc}
    \toprule
      \multirow{2}{*}{\textbf{Model}}   & \multicolumn{2}{c}{\textbf{ObjHal}} & \multicolumn{2}{c}{\textbf{MMHal}} \\

      \cmidrule(lr){2-3} \cmidrule(lr){4-5}
      
      & Resp.$\downarrow$  & Mention$\downarrow$  & Info.$\uparrow$ & Resp.$\downarrow$\\
    \midrule
      LLaVA & 63.0 & 29.5 & 31.9 & 70.8 \\
      RLHF-V & 12.2 & 7.5 & 40.0 & 52.1\\
    \midrule
    \rowcolor{mygray} POPEN & \textbf{9.2} & \textbf{4.9} & \textbf{42.1} & \textbf{47.9}\\
   \bottomrule
    \end{tabular}
    }
    \caption{Results on hallucination benchmarks.}
    \label{table_hal_bench}
\end{table}

\subsection{Hallucination Mitigation on MUSE} Some previous works have explored ways to mitigate hallucinations in LVLMs. We compare our method with these approaches on the MUSE validation set, and the results are presented in Table \ref{tabel_mitigation}. Although these previous methods can alleviate hallucination to some extent compared to the baseline, our approach significantly outperforms them, with substantially reduced $C_{S}$ and $C_{I}$ metrics. Moreover, due to the lack of segmentation-specific designs in previous methods, they fail to achieve significant improvements in segmentation accuracy. In contrast, benefiting from preference-based optimization and ensemble techniques specifically designed for segmentation, our method, POPEN, greatly enhances segmentation metrics including gIoU and cIoU, demonstrating the superiority of our approach.

\subsection{Improvement of Target Localization}
As indicated in Sec.3.2 of the main paper, during the first half of finetuning, we collect perturbed images with varying localization accuracy as preference data $\mathcal{P}_{s}$ for optimization. To further validate the effectiveness of this method, we conduct a quantitative comparison of target localization precision between models finetuned using randomly generated $\mathcal{P}_{s}$ and those finetuned with $\mathcal{P}_{s}$ generated by our method. Specifically, for each object mask generated by the model, we calculate its IoU $IoU_{g}$ with the ground truth object mask, as well as its maximum IoU $IoU_{o}$ with other objects in the image (provided by SAM) beyond the ground truth object. We then define a mask as having target localization error if $IoU_{g}<0.75$ and $IoU_{o}>0.25$, and we compute the proportion $p$ of such wrongly-located objects among all objects in the MUSE validation set. On this metric $p$, the model finetuned with randomly generated $\mathcal{P}_{s}$ achieves a score of 23.3\%, while the model finetuned with $\mathcal{P}_{s}$ generated by our method reduces this to 6.5\%. This demonstrates the significant improvement on the model’s target localization capability brought by our method.

\subsection{Results on Stronger LVLMs}
In addition to LLaVA, several more advanced LVLMs have been proposed recently, offering better performance and reduced hallucination for different tasks. We further evaluate the effectiveness of POPEN when integrated with these stronger LVLMs. As shown in Table \ref{qwen_results}, replacing LLaVA with a stronger Qwen2-VL-7B \cite{wang2024qwen2} does lead to some performance improvement, including better segmentation quality and reduced hallucination.  However, even when using a weaker LVLM, LLaVA + POPEN still outperforms Qwen2-VL + PixelLM (with a stronger LVLM), demonstrating that simply relying on a stronger LVLM is not sufficient; while using our novel methods in POPEN can yield larger improvements. Additionally, the notable advantage of Qwen2-VL + POPEN over Qwen2-VL + PixelLM further highlights the ability of our method to improve performance even under a stronger LVLM, demonstrating its high effectiveness and generalizability for different base models.

\begin{table}[t]
    \centering
    \setlength\tabcolsep{10pt}
    \resizebox{0.87\linewidth}{!}{
    \begin{tabular}{l|cccc}
        \toprule
        Method & gIoU $\uparrow$ & cIoU $\uparrow$ & $C_{S}$ $\downarrow$ & $C_{I}$ $\downarrow$\\
        \toprule
        Baseline (PixelLM) & 41.9 & 48.9 & 22.0 & 9.8\\
        \midrule
        OPERA \cite{huang2024opera} & 42.3 & 49.5 & 15.0 & 6.8\\
        VCD \cite{leng2024mitigating} & 42.3 & 49.3 & 16.9 & 7.3\\
        HALC \cite{chen2024halc} & 42.2 & 49.6 & 14.2 & 6.7\\
         \midrule
         \rowcolor{mygray} POPEN \cite{chen2024halc} & \textbf{45.4} & \textbf{55.2} & \textbf{9.3} & \textbf{4.3}\\
         \bottomrule
    \end{tabular}}
     \caption{Comp with other hallucination mitigation methods.} 
    \label{tabel_mitigation}
\end{table}

\begin{table}[t]
    \centering
    \setlength\tabcolsep{5pt}
    \resizebox{0.85\linewidth}{!}{
    \begin{tabular}{l|cccc}
        \toprule
        Method & gIoU $\uparrow$ & cIoU $\uparrow$ & $C_{S}$ $\downarrow$ & $C_{I}$ $\downarrow$\\
        \midrule
         LLaVA + PixelLM & 41.9 & 48.9 & 22.0 & 9.8\\
         Qwen2-VL + PixelLM &  42.8 & 50.5 & 15.2 & 7.3\\
         \midrule
         LLaVA + POPEN &  45.4 & 55.2 & 9.3 & 4.3\\
         \rowcolor{mygray} Qwen2-VL + POPEN &  \textbf{46.1} & \textbf{56.4} & \textbf{8.1} & \textbf{3.8}\\
         \bottomrule
    \end{tabular}}
     \caption{Effectiveness on the stronger LVLM Qwen2-VL.} 
    \label{qwen_results}
\end{table}

\subsection{Ablation Study of Hyperparameters}
In our method, the hyperparameters $\beta_{t}$ in Eq.2 and $\lambda$ in Eq.3 of the main paper follow the same settings as RLHF-V \cite{yu2024rlhf}. Therefore, we primarily focus on validating the remaining hyperparameters in our approach, including $\beta_{s}$ in Eq.4 of the main paper, and the number $K$ of generated responses in the preference-based ensemble method. Both the text metric $C_{S}$ and segmentation metric cIoU are reported. (higher cIoU and lower $C_{S}$ are better.)\\

\begin{figure}[t]
    \centering
    \includegraphics[width=1\linewidth]{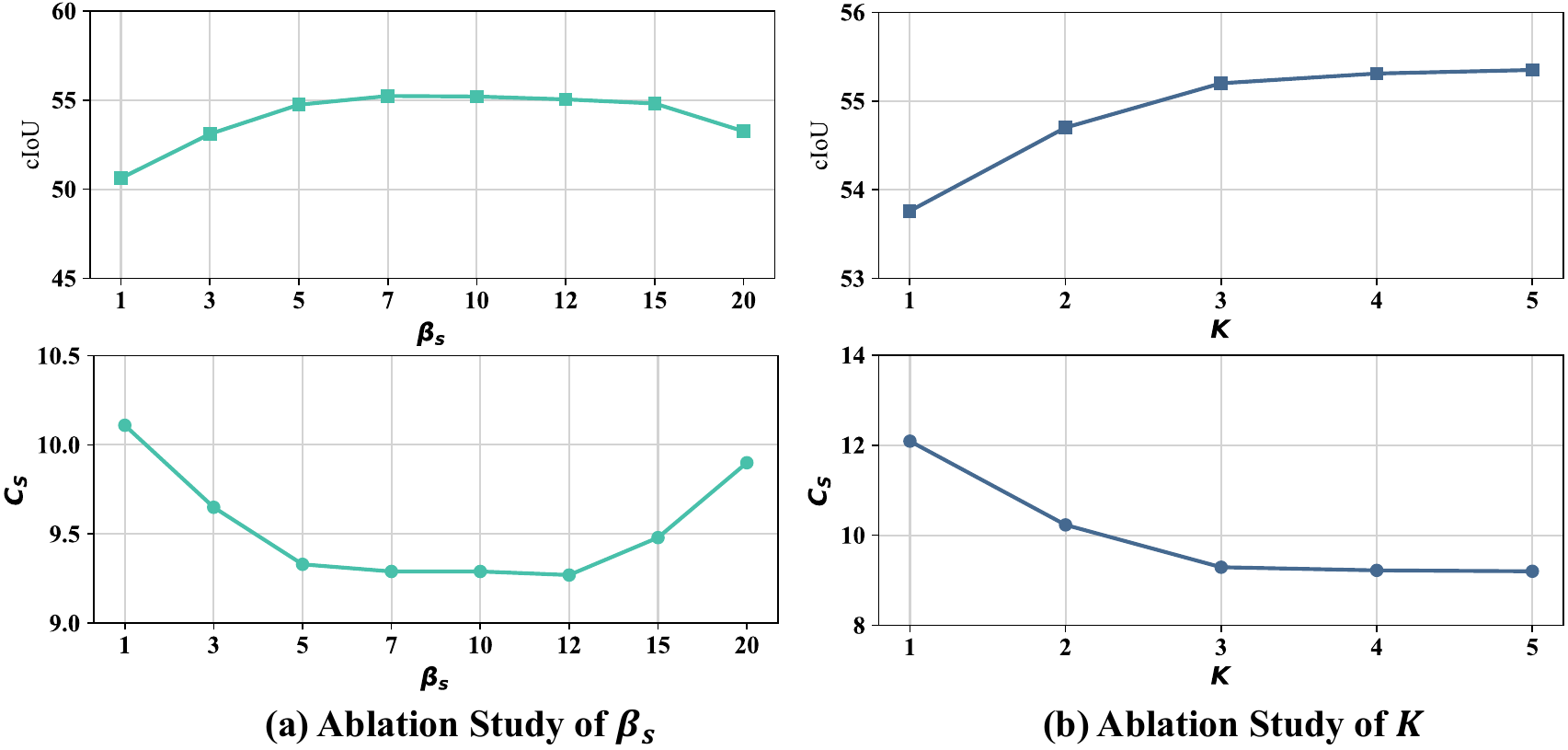}
    \caption{Ablation study of hyperparameters $\beta_{s}$ and $K$ on the cIoU metric (the first row) and $C_{S}$ metric (the second row). Higher cIoU and lower $C_{S}$ indicate better results.}
    \vspace{-0.5\baselineskip}
    \label{ablation_fig}
\end{figure}

\noindent \textbf{Ablation Study of $\beta_{s}$.} $\beta_{s}$ is a scaling factor used in Eq.4 of the main paper to compute the segmentation preference loss. As shown in Figure \ref{ablation_fig}(a), overly small or large values of $\beta_{s}$ can lead to performance degradation. However, the performance can remain consistently stable when $5<\beta_{s}<15$, demonstrating the robustness of our method to the choice of $\beta_{s}$.\\

\noindent \textbf{Ablation Study of $K$.} We further evaluate the impact of $K$, with the results presented in Figure \ref{ablation_fig}(b). Increasing $K$ enhances performance, as the quality of the refined results can be improved through the fusion of more responses. However, performance plateaus when $K>3$, with only marginal gains observed upon further increases. Therefore, we select $K=3$ as our setting.

\subsection{Further Ablation of Curriculum Collection} As detailed in Sec.3.2 of the main paper, we employ a curriculum method for obtaining segmentation embedding preference data $\mathcal{P}_{s}$, collecting different types of $\mathcal{P}_{s}$ for the first and second halves of finetuning. In Table 5 of the main paper, we conduct an ablation study to compare our method with random collection or using the same strategy throughout both halves of the finetuning process. In this Supp, we further compare with a hybrid approach, where samples collected using the first-half strategy and those collected using the second-half strategy are both used throughout the entire finetuning process. As shown in Table \ref{supp_table_ablation_collection}, this hybrid approach outperforms random collection but remains significantly inferior to our curriculum-based method. This demonstrates the importance of sequentially learning fundamental and advanced skills in our proposed method.

\subsection{More Qualitative Comparison}
\begin{figure*}[t]
    \centering
    \vspace{-0.5\baselineskip}
    \includegraphics[width=0.95\linewidth]{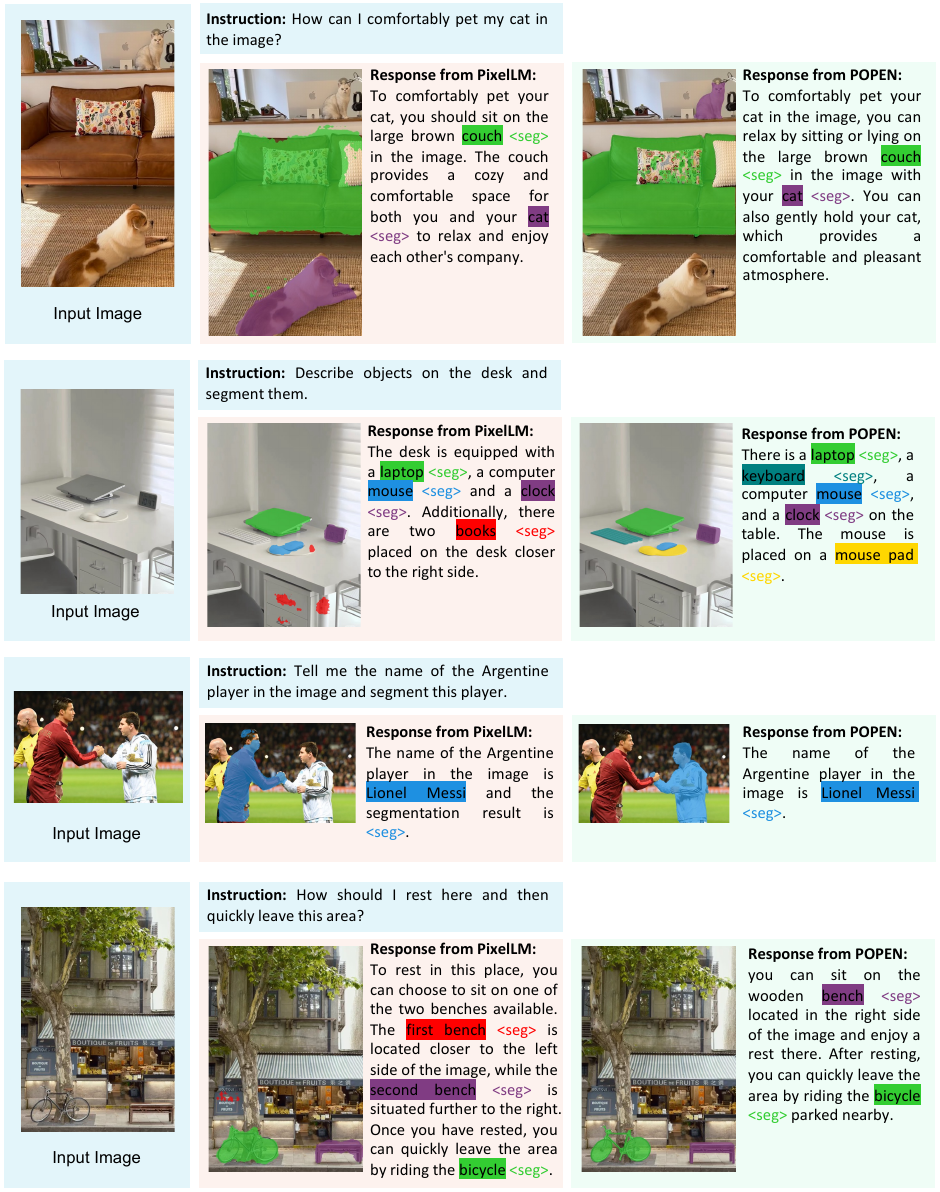}
    \vspace{-0.5\baselineskip}
    \caption{More comparative examples of text responses and segmentation results between PixelLM and our POPEN.}
    \vspace{-0.5\baselineskip}
    \label{supp_example_fig}
\end{figure*}

In Figure \ref{supp_example_fig}, we present more examples comparing text responses and segmentation results between our POPEN and PixelLM \cite{ren2024pixellm}. In these examples, PixelLM suffers from serious hallucinations, generating objects in its text responses that do not exist within the images, such as the ``books'' in the second example and ``bench in the left side" in the fourth example. Furthermore, the segmentation accuracy is suboptimal, with coarse results in the object boundary regions and even wrong localization of the target objects (such as the segmentation for ``cat" in the first example and ``Lionel Messi" in the third example). By employing the proposed preference-based optimization and ensemble methods, our POPEN achieves significantly improved results, effectively mitigating hallucination in text responses and enhancing segmentation accuracy. These comparative results demonstrate the high effectiveness and advantage of our method compared to PixelLM.

\begin{table}[t]
    \centering
    \setlength\tabcolsep{8pt}
    \resizebox{1\linewidth}{!}{
    \begin{tabular}{l|cccc}
        \toprule
        Collection Method & gIoU $\uparrow$ & cIoU $\uparrow$ & $C_{S}$ $\downarrow$ & $C_{I}$ $\downarrow$\\
        \midrule
        Curriculum Collection & 45.42 & 55.20 & 9.29 & 4.31\\
        \midrule
         Random & 43.06 & 51.35 & 9.78 & 4.62\\
         Hybrid Approach & 44.19 & 53.75 & 9.48 & 4.55\\
         \bottomrule
    \end{tabular}}
     \caption{Effectiveness of different methods for segmentation preference data collection. ``Hybrid approach" refers to samples collected using the first-half strategy and those collected using the second-half strategy are both used throughout the entire finetuning process.}
     \vspace{-0.5\baselineskip}
    \label{supp_table_ablation_collection}
\end{table}

\end{document}